# Interval-Based Decisions for Reasoning Systems

Ronald P. Loui    Jerome A. Feldman    Henry G. E. Kyburg, Jr.

Departments of Computer Science and Philosophy
University of Rochester
Rochester, NY 14627

Abstract and Claims.

This essay tries to expound a conception of interval measures that permits a particular approach to partial ignorance decision problems. The virtue of this approach for artificial reasoning systems is that the following questions become moot: 1. which secondary criterion to apply after maximizing expected utility, and 2. how much indeterminacy to represent. The cost of the approach is the need for explicit epistemological foundations: for instance, *a rule of acceptance* with a parameter that allows various attitudes toward error. Note that epistemological foundations are already desirable for independent reasons.

The development is as follows: 1. probability intervals are useful and natural in A.I. systems; 2. wide intervals avoid error, but are useless in some risk-sensitive decision-making; 3. yet one may obtain narrower, or otherwise decisive intervals with a more relaxed attitude toward error; 4. if bodies of knowledge can be ordered by their attitude to error, one should perform the decision analysis with the acceptable body of knowledge that allows the least error, of those that are useful. The resulting behavior differs from that of a Bayesian probabilist because in the proposal, 5. intervals based on successive bodies of knowledge are not always nested; 6. the use of a probability for a particular decision does not require commitment to the probability for credence; and 7. there may be no acceptable body of knowledge that is useful; hence, sometimes no decision is mandated.

I. Interval Measures.

By now, the use of an interval measure is regarded highly for probability judgements in reasoning systems. Researchers selecting formalisms for quantifying belief have all recognized the virtues of (partial)[1] indeterminacy in probability judgement ([Bar81], [GLF81], [Dil82], [Low82], [WeH82], [Qui83], [Wes83], [Gin84], [LuS84], [Str84], etc.).

Intervals allow varying degrees of commitment in probability assertion. At the extremes, $P(A) = [0, 1]$ is uncommitted, while $P(A) = [.76, .76]$ is consummate. Some have argued that indeterminacy captures "pre-systematic" notions of belief and disbelief [Sha76], [Lev80a].[2] Since $0 \leq \inf P(A) + \inf P(\sim A) \leq 1$, the agent can assign zero belief to a proposition even though he is not certain that it is false. Indeterminacy is useful to the subjectivist when eliciting bounds on probabilities (especially from equivocating experts), and to the empiricist for expressing the Neyman-Pearson confidence results of population sampling.

Intervalism is also natural in detachment. When $\%(A, Q) = 1 - \delta$, then $\%(A, P\&Q) = [1 - \epsilon - \delta, 1]$. If probabilities are based on direct inference from the class $A$, the probability of "$Px \& Qx$" for some $x \in A$ would be an interval, despite having started with probabilities that were points (see [TSD83], [Che83], and [Nil84]).

Many advocates of interval belief measures in A.I. link their arguments to Shafer's interpretation [Sha76] of Dempster's inference system [Dem68]. Shafer's theory is claimed to provide a valuable representation of intervals (via mass functions), and a simple, consistent approach to resolving apparent disputes when combining evidence (via Dempster's rule). These claims are evaluated elsewhere [Kyb85], [Lev80a], [Zad79]. Shafer's theory is not unique in its ability to cope with disagreeing evidence; indeed, a system of belief would be impoverished if it made no provisions for disagreement (see Levi's remarks [Lev80a]; also, there are indeterminate systems due to Levi, Smith, Schick, Good, and Kyburg). Further, Dempster's rule for combining evidence is relatively

193

presumptuous as a form of conditionalization [Dem68], [Lev80b], [Kyb85].

Putting aside the prospects for Dempster's rule, we are left with these indeterminate probabilities, and with an ensuing decision problem. Barnett [Bar81] and Lowrance [Low82] have both suggested that a research goal should be a fully developed decision theory based on interval measures.

Luce and Raiffa call decision problems with indeterminate probabilities "partial ignorance" problems, and earlier work on partial ignorance is discussed in [Lou85]. Wesley, Lowrance, and Garvey [WLG84] offer a candidate theory that is for use with Shaferian beliefs and that ignores risk; it has been discussed elsewhere [Lou84].

II. Estimation and Decision.

With interval probabilities or interval utilties, expected utilities are intervals. If interval probabilities are narrow (or otherwise fortuitous) there is no problem: expected utility intervals can be ordered in the natural way (see below), and the best act identified. In a 1:1 lottery that depends on the outcome of a coin toss, if $P(heads)$ is $[.7, .8]$ the decision should be clear via the obvious ordering; if it is $[.3, .8]$, the decision may not be clear. The decision may also not be clear if the interval is narrow, but unfortuitous, e.g., $[.49, .52]$.

If the maximization of expected utility (MEU) is the sole solution criterion, there may be no defensible ordering of the utility intervals that identifies a best act. Of course, MEU with point-probabilities can be ambiguous too. This latter ambiguity is often tolerated: if two acts have the exact same expected utility, the sameness of utility is supposed to reflect indifference. But ambiguity with interval probabilities may not be tolerable because intervals often model ignorance, not indifference. It is not the case that the two acts *couldn't* be ordered in a relevant and accountable way. Rather, not enough is known to order them.

There are two problems here. First, there is the estimation problem: what should be the degree of certainty attributed to a proposition? Second, there is the decision problem: which act should be chosen among available acts, when the agent is not indifferent about them all? In the estimation problem, error is avoided by using intervals. In the decision problem, ambiguity is avoided by eschewing intervals. In order to solve both problems simultaneously, there must be some compromise.

III. Secondary Criterion Solutions.

Let $\Pi$ be the largest set of probability distributions satisfying all of the interval constraints. Calculating expected utilities in the usual way, for act $a_\lambda$, in the presence of uncertainties $E_i$:
$$u_k(a_\lambda) = \Sigma \{ P_k(E_i|a_\lambda) u(\langle E_i, a_\lambda \rangle) \}$$
$$\forall E_i;$$
$$U(a_\lambda) = \{ u_k(a_\lambda) : P_k \in \Pi \};$$
$$u(a_\lambda) = [inf\, U(a_\lambda), sup\, U(a_\lambda)].$$

The natural way to (partial-) order acts with indeterminate utilities is by dominance: $a_1 > a_2$ iff $inf\, U(a_1) > sup\, U(a_2)$. If there is a unique maximal element in the order, $a^*$, then the decision problem is solved. The probabilities, though individually indeterminate, are nevertheless collectively *decisive*. But in general, there will be some set of maxima, $\{a_i\}$.

Some authors ([Hur51], [Goo83], [Fis65], [Lev80b]) suggest that $a^*$ can be identified in the maximal set by one of the so-called weaker methods: maximin, min-regret, or lexi-min methods. These are the methods recommended for decision problems under uncertainty, and their common character that is crucial here is that they make no use of probability judgement. Presumably the probability information has been milked for all it is worth, under the primary method of MEU, and secondary methods will finish the job of identifying $a^*$. Unless there is an unforseen equality of point-valued utilities, the weaker method guarantees identifying a unique act. The weaker method *could* have been applied in the first place but for its admitted weakness. It is considered weak precisely because it ignores probability judgement. It is employed secondarily precisely because it presupposes that probability judgement will be of no further use, which is exactly the case among the maximal set after the application of MEU.

For programmed systems, there is still the problem of choosing one from among the various secondary criteria. Clearly there are situations in which maximin is inappropriate, and similarly for min-regret, for optimism-pessimism, etc. Supervaluations would be cautious, but impotent; taking the most popular mandate among various criteria would be ad hoc. One could attempt to discriminate those situations in which one method applies and others do not, but no such attempt has been successful.

IV. A Different Proposal.

Here is an entirely different way of solving partial ignorance problems. If MEU with the given probability intervals is indecisive, MEU can be retained and the probability intervals refined. In the Bayesian tradition, refinement of intervals is done subjectively, with no additional empirical information. In the Neyman-Pearson tradition, refinement is done objectively and requires additional empirical information. In either case, refinement further determines probabilities. An automatic reasoning system may be required to be objective, may not have recourse to additional information, and may require the preservation of indeterminacy. Fortunately, it's possible to refine intervals objectively, with no additional empirical information, and without losing the indeterminacy of probabilities. This latter possibility is presented more carefully in [Lou85].

Let credal state be described not by one set of feasible distributions, $\Pi$, but by a sequence of sets, $\langle\Pi_i\rangle$. Each $\Pi$ is based on a body of knowledge formed with some quantifiable attitude toward error (so there is a companion sequence, $\langle K_i\rangle$, where each body of knowledge, K, has an integer index and a real error). Successive K's are more informative, but predecessors are less prone to error. Each indeterminate expected utility calculation is done with respect to one element in the $\Pi$-sequence, but the whole $\Pi$-sequence constitutes the credal state. Therefore, different indeterminate probabilities, with different maximal sets, can be consulted for the purposes of decision without changing the indeterminacy of the credal state.

This representation finesses the question of how narrow intervals ought to be. Imagine the expert who first reports the interval is [.3, .7], but can be coaxed into reporting to the more useful [.35, .65]. Which interval gets represented? In this proposal, both should be represented. Intervals should be as narrow as permitted given the magnitude of error ($a$) associated with the body of knowledge on which the intervals are based, $K_{(a)}$. They will be [0, 1] in $\Pi_0$. They may be degenerate in the very late $\Pi$'s. And they should be *variously* narrow (though not necessarily nested) in between.

In practice, this proposal requires additional represented information, or additional inference rules and epistemological assumptions. It may be possible simply to assert and to represent both sequences, $\langle\Pi_i\rangle$ and $\langle K_i\rangle$. But more likely, $\Pi$'s will have to be generated from K's, and successive K's from some initial base, $K_{init}$. A combination of the two methods, generation and assertion, is convenient.

Generating $\Pi$'s from K's requires the adoption of some theory of probability. It could be as simple as taking statements in K to be constraints on distributions, or conditionalizing some prior on the contents of K, or it could be some theory of frequency-based or chance-based direct inference.

Generating $succ(K)$ amounts to making additional assumptions. It could be done in a number of ways: one possibility is to use an acceptance rule (see also below, on "higher-order" probabilities). Such a rule would describe when a statement is acceptable and would thus determine to which K's it belongs. If the rule is based on probabilities relative to $K_{init}$ for instance, then $A$ belongs to all those successor K's, $K_j$, such that $1 - P(A \mid K_{init})$ is less than the error associated with $K_j$. A different probabilistic rule would take $succ(K)$ to be $K \cup \{A\}$, where $A$ is the next most probable statement relative to K, of statements of some special form. Note that with these rules, the K-sequence is nested. Acceptance rules in the literature are more elaborate. See [Kyb70] for additional acceptance rules and their evaluation.

Decision-making amounts to exploring the $\Pi$-sequence in best-first order until either (1) the maximal set under $\Pi$ is a singleton, in which case the problem is solved; or (2) $\Pi$ is a singleton, which leaves the standard decision problem under risk; or (3) the error associated with $\Pi$ is intolerably large for this decision problem, in which case MEU with no acceptable set of assumptions can legislate unambiguously.

The reasonableness of this proposal depends on whether there is independent reason to use intervals of a particular width. It may be that epistemological considerations require that certain intervals be used: e.g., the narrowest intervals at .95 confidence [Kyb85]. But if not, if confidence levels .94 and .96 are also useable, then decision analysis might as well proceed with intervals that are decisive, rather than with those that are indecisive. There is no reason to avoid tolerable error if doing so results in uninformative analysis.[3] If the MEU calculation is not satisfactory under the assumptions held, it could be that the agent has not assumed enough. The analysis should then be founded on an augmented set of assumptions.

Conversely, there is no reason to invite error in the analysis if the analysis is already sufficiently informative. So of the many $\Pi$'s that are decisive, the one that is least prone to error has epistemic priority. The augmented set of assumptions should be the next-least in order of presumptiveness. No more assumptions should be made than are necessary for decision.

Consider the claim that rational commitment ceases with the restriction to the maximal set, or that the agent must sometimes suspend judgement when the set of maxima is not a singleton. Lopes, voicing a common intuition, quips that suspending judgement among choices with overlapping expected utility ranges is no more defensible than suspending judgement among choices with overlapping outcome ranges [Lop83]. Lopes' remark is forceful precisely because it points out the arbitrariness of interval width. Why invite error by using intervals narrower than [0, 1]? Because [0, 1] intervals are not satisfactorily informative. But if the maximal set under narrower intervals is unsatisfactory, and the limit of tolerable potential error has not been reached, why not use still narrower intervals? One is already willing to forego certainty, and the amount of certainty one is willing to forego depends on the other desiderata, including decisiveness.

We still avoid error by using indeterminacy: we retain the early elements in the $\langle \Pi_i \rangle$ sequence, rather than settling immediately on the most specific element (or of some P, s.t. $P \in \Pi_i$ for all i). There may not be a most specific element in the sequence (this is explored in example C, below). And there may be genuine instances in which no substantiable set of assumptions legislates a unique decision or formulates a standard risk problem. In such cases, indeterminacy is required to indicate ignorance, or if either is possible, the need for more sampling, or for suspension of judgement.

For example, consider a probabilistic acceptance rule: statements are accepted in $K_{(\alpha)}$ when their probability relative to $K_{init}$ exceeds $1 - \alpha$. For a decision problem where the maximum ratio of odds is $w$, it would be pointless to perform an MEU analysis in some $K_{(\alpha)}$ where $\alpha \geq 1 - w$. If the lottery pays 20:1, $w = .95$. If all $\Pi$'s based on less error than .05 are indecisive, no decision is legislated (see [Kyb85] and [Lou85] for discussion).

V. Examples and Contrasts.

We discuss the following decision problem.[4] Upon finding a berry, the agent has to decide whether to eat it ($a_1$), or not to eat it ($a_2$). If it is eaten, it matters whether or not it was a good berry ($G$). If it is not eaten, it matters whether or not the agent later gets hungry ($H$). Let $u(\langle a_1, G \rangle) = 10$; $u(\langle a_1, \sim G \rangle) = -30$; $u(\langle a_2, H \rangle) = -10$; and $u(\langle a_2, \sim H \rangle) = 0$.

A. *lower level confidence intervals.*

Suppose the probability reports for $G$ and for $H$ are based on Clopper-Pearson intervals. Of 4 berries eaten, 4 were good. On 14 excursions of this kind, the agent got hungry (without eating) 3 times. At .99 confidence, $P(G) = [.35, 1]$ and $P(H) = [0, .55]$. So $u(a_1) = [-16.8, 10]$; $u(a_2) = [-5.5, 0]$. The maximal set is $\{a_1, a_2\}$. But at the confidence level .75, $P(G) = [.75, 1]$ and $P(H) = [.15, .3]$. So $u(a_1) = [0, 10]$; $u(a_2) = [-3, -1.5]$. $a_1 > a_2$. $a_1$ is uniquely maximal. Note that if $a_1$ and $a_2$ had been ranked by utility midpoints at .99, $mp_1 = -3.4$; $mp_2 = -2.75$; one would have concluded contrarily that $a_2 > a_1$!

B. *direct inference and probabilistic acceptance rule.*

Suppose %(*berries, good*) = [.3, .8] and %(*excursions, get-hungry*) = [0, 1] and %(*soft berries, good*) = [.84, .88]. Presumably this is accepted based on sampling with, say, at least .999 confidence. If $P(G)$ is based on the [.3, .8] interval, both $a_1$ and $a_2$ are maximal. The decisive [.84, .88] interval can't be used for $P(G)$ unless it is *accepted* that the berry is soft. Even if there is independent reason to believe $P($*this berry* $\in$ *soft berries*$) = .999$, the probability of $G$ would be [.3, .8]. It's natural to consider the acceptance of "*this berry* $\in$ *soft berries*". This allows direct inference: $P(G)$ must be [.84, .88] if this is all that is known.[5] Decision to do $a_1$ is based on dominance with the narrower interval.

C. *convex Bayesian vs. Savage's Bayesian.*

A Bayesian who considers all the distributions in a closed convex set can accept different constraints on this set at different levels of acceptance (cf. [Lev80b]). Typical constraints could be conditions (as in example B), or bounds on marginal probabilities (as in example A). Additional knowledge can lead to additional constraints, which can decrease membership in $\Pi$ and so are more informative (though additional knowledge does not always lead to additional constraints: sometimes it can invalidate a constraint). Some constraints may not be as warranted as others, and their use introduces more possibility of error. If the set is indecisive, try the MEU analysis with the next set of constraints.

Savage would have the agent settle on the most specific set (if there is one), and eliminate the excess indeterminacy of the preceding sets. If all the sets are nested (for all $i > j$, $\Pi_j \supseteq \Pi_i$), there is no difference between the decisions made by this convex Bayesian and by Savage's Bayesian.

But sets are not nested. The most obvious source of non-nesting is due to conditionalization.[6] Suppose $\Pi_1$ is based on acceptance so stringent that probabilities are conditional only on $A$. $\Pi_2$ takes both $A$ and $B$ as conditions; $B$ is acceptable as a condition at this level (perhaps $B$ is treated by Jeffrey's rule in $\Pi_1$; it doesn't matter here). Then there's no reason for $\Pi_2$ to be a subset of $\Pi_1$.

Let $A$ entail $\sim H$; $P(G \mid A) = [.6, .8]$; $P(G \mid A, B) = [.3, .4]$. Intuitively, $A$ might be the conjunction "*just ate & the berry looks good*" while $B$ might be "*the lighting is misleading*". $\Pi_1$, with $P(G) = [.6, .8]$, indicates both $a_1$ (eating) and $a_2$ (not eating) as maximal. $\Pi_2$ mandates $a_2$, with $P(G) = [.3, .4]$, which is not a sub-interval of [.6, .8]. Now suppose the next decision involving $P(G)$ is a 1:1 lottery. $\Pi_1$ mandates entering the lottery, and because $\Pi_1$ is decisive and epistemically prior, $\Pi_2$ is ignored. Savage would continue to use $P(G)$ from $\Pi_2$, and would avoid the lottery.

So preservation of the "excess" indeterminacy is necessary despite temporary refinement for the purposes of the current decision.

D. *Shaferian discounting.*

It's tempting to consider Shafer's discounting parameter to generate successive $\Pi$'s.

The belief with mass $m(G) = .7$ and $m(\sim G) = .3$ is to be combined with a belief $m(G) = .6$; $m(\sim G) = .4$ based on a new, independent source. The latter's impact is to be discounted by some amount $r$. Let $\sim H$ be accepted. If $r < .23$, then $P(G) > .75$, and $a_1 = a^*$; otherwise $a_2 = a^*$. Note that for any value of $r$ here, the resulting probability of $G$ is determinate.

Are some values of $r$ more cautious than others? If $r$ is large, the informative impact of the second belief is lessened, and it is combined with caution. But a cautious attitude toward the new belief is not necessarily a cautious attitude toward the possibility of error, unless the new belief is the only possible source of error. When conditions were not accepted in example C, it was because they were relatively uncertain, not because they were new. Here, it may be that the full weight of the new belief is required to avoid error. It would be erroneous, for instance, to ignore the new belief completely. The parameter $r$ here is being used like Carnap's $\lambda$. There is no epistemic relation given between $r$ and error, hence, no priority of one solution over the other.

Perhaps caution should be reflected by discounting both belief functions. This begs the question, in what proportion should they be discounted? If there are two parameters that can be varied, the $\Pi$'s generated will be only partially ordered.

It may be possible to use Shafer's formalism to generate the $\Pi$-sequence, but its use would require more argument.

## VI. Epistemological Considerations.
### A. *On Revisions of the Knowledge Base.*

A behavioral interpretation of probability suggests the identification of $a^*$ as additional evidence about probability judgement. Whatever the means of $a^*$'s identification, there is a set, $\Psi$, of admissible probability distributions, according to each of which, $a^*$ is the unique maximum by MEU means alone. Behaviorists hold that once $a^*$ is identified, the agent's credal state contracts to the more precise $\Pi'$, the intersection of $\Psi$ and $\Pi$, at least as a description appropriate at the time of decision. Presumably, if there is no subsequent revision, the more precise description of past state continues to describe the current state. If this is right, then credal state depends on the decisions made. Faced with a different decision structure, $a^*$, hence $\Psi$, and finally credal state, might have been different.

Upon each decision, the agent must be consistent, in this behaviorally strong sense. Decisions always reveal credal state and always do so through MEU.

There are enormous implications of this revelation-through-behavior stance for the management of knowledge bases. No matter how tentative the decision, and whatever its content or manner of selection, the knowledge base must represent only the distributions that are MEU-admissible for that decision. If only a single distribution is MEU-admissible, then that distribution specifies the new state of the program's belief. And this has been done with the addition of no relevant empirical knowledge! All that distinguishes the new state from the old is the actualization of one particular problem structure, among the many that could have been faced.

If the interpretation of probability is subjective as well as behavioral, the agent or reasoning system can spuriously return to the more permissive credal state, $\Pi$. But if this is to be a rule for revision, there seems no point in making the contraction. If it is not a rule, then there is still the onerous possibility of spurious change to some other credal state, and worse, the possibility of no change whatsoever after contraction.

Either course violates legitimate counterfactual intuitions pertaining to the past decision. Suppose the secondary method is always a tournament of coin-flipping. Upon the last toss of *heads*, $a_2$ is chosen, and $\Pi'$ is obtained from $\Pi$ by the deletion of all distributions that do not mandate $a_2$. Thus, it no longer is the case that "had the toss been tails, $a_1$ would have been mandated," though we quite reasonably take such to have been the case.

Starr [Sta66] suggests a normative criterion for identifying the optimal act when $\Pi$ is not a singleton. Suppose the distributions in $\Pi$ can be parameterized by some $\theta$. Suppose also that the set of parameter values $\Theta$, corresponding to the $\Pi$ distributions, is measured by an additive indifference "prior". So subsets of $\Pi$ are also measured. Consider various acts. An act is mandated by each of its MEU-admissible distributions, which collectively form some subset $\pi \subseteq \Pi$. Starr's criterion chooses the act with the $\pi$ that maximizes the measure (i.e., that has the greatest number of feasible admissible distributions).

Starr's criterion is a prescription for decision, not for the adoption of a narrower credal state. Behaviorists would contract to $\pi$.

Whatever the behaviorist arguments, the revelation of credal state through decisions and MEU is unattractive in A.I. A system's probability estimates are based on objective analysis of samples, or on the opinions of experts, not on the future decision problems to be faced by the system.

B. *On Higher-Order Probabilities.*

Some Bayesians intuit the existence of "higher order" probabilities (e.g., [Goo83]). These would be probability distributions on probability distributions, formalized perhaps, like the indifference "prior" in Starr's criterion.

If one approves of and has access to such measures, then acceptance can be based on the measure. For instance, successive $\Pi$'s could be generated by eliminating the next-least probable members of the previous $\Pi$. This strategy leads to nested $\Pi$'s; all decisions would be those mandated by the distribution with the greatest higher-order-measure. It would not, in general, be the same as taking an expectation over the expected utility intervals, and ranking the resulting real-values:
$$u(a_\lambda) = \Sigma\{\Sigma\,[P_k(E_i|a_\lambda)u(<E_i, a_\lambda>)]M(P_k)\}$$
$$\forall k\,\forall i,$$
where $M$ is the higher-order measure.

Perhaps the expectation is appropriate if there is such a measure. However, one should have misgivings about the identification of these measures.

There may be uncertainty about the higher-order measure, reflected in some still higher measure. This induces a hierarchy of measures. Presumably the height of the hierarchy is finite. There must be, at some high order, either a determinate measure, or else unmeasured indeterminacy. If the former, then one should be suspicious about the source of a determinate higher-order-measure: why is the probability of a distribution certain, but the distribution uncertain? The higher-level is not inherently more robust (note that the order of the sums can be reversed). Just as a small error in a probability can change a decision, so can a small error in a higher-order probability change a decision.

If on the other hand there is unmeasured indeterminacy, the expected-expected utilities will be intervals. This is essentially no different from the interval expected utilities from indeterminate zero-order probabilities.

So acceptance can be conceptually related to higher order probabilities, but is not immediately subsumed or improved by them.

VII. Conclusion.

A.I. systems that use interval judgements must sometimes solve partial ignorance decision problems. There are now two approaches. Maximizing expected utility can be followed by maximin, or some other secondary criterion. Alternatively, additional assumptions can be made that change probabilities, temporarily, so that maximizing expected utility is sufficient. This paper has discussed how to implement the latter approach. Assumptions are accepted in an order that tries to avoid error, and they are accepted only temporarily, for the purposes of decision.

There is still the problem of choosing an acceptance rule, which iteratively generates the next-best assumption. This choice requires considerably more epistemological reflection.

## VIII. Notes.

[1] By *indeterminacy*, we will mean indeterminacy broadly construed: potential indeterminacy, including the judgements Prob(A) ∈ [0, 1] (complete) and Prob(A) ∈ [.1, .1] (degenerate), Prob(A) ∈ [.3, .7] (bounded), and [Prob(A) = .4 or Prob(A) = .8] (disjunctive).

[2] Some have charged that the specification of an interval requires two numbers rather than one; hence, it requires more information. That's silly. Given that some quantity p is in fact 0.67, it follows that p is in the interval [0.34, 0.97]. Furthermore, in a very natural canonical form, namely, the number of hyper-planar constraints required in the space of all probability distributions; the information (number of constraints) in interval reports of a particular probability is less than the information in point reports. Information measures are dependent on canonical form, hence can be misleading.

Intervals are chosen because they offer robust behavior. If practice shows that they are not robust enough, that endpoints matter critically, then future investigators can feel free to use a formalism with indeterminate upper and lower bounds, or with fuzzy sets. Surely one would not revert to point probabilities because they contain "less information."

[3] Here, we've taken informativeness w.r.t. decision to be singularity of $\Pi$ or singularity of the maximal set. Other interpretations of "informative" are possible (such as any restriction of the maximal set to decisions which cannot differ in outcome more than $\epsilon$). These lead to different decision theories.

Also note that in [Lou85], the amount of tolerable error is addressed (see the discussion of D-meaningful corpora).

[4] We call this the problem of Jerry's Berries.

[5] We've appealed to the epistemological conception of probability here. If explicit statement of chances is required, the example can be changed

[6] It's also possible to violate nesting when constraints are ordered jointly, and not all constraints are compatible. So if $c_1, \ldots, c_4$ are constraints on $\Pi$'s, $\Pi_1$ may be delimited by $\{c_1\}$, and $\Pi_2$ by $\{c_2\}$, and $\Pi_3$ by $\{c_4\}$ before $\Pi_3$ by $\{c_1, c_2\}$. $\Pi_5$ may be delimited by $\{c_1, c_4\}$, where $\{c_1, c_2, c_4\}$ is over-determining. If constraints are accepted (rather than knowledge that generates constraint), and acceptance is purely probabilistic, then this kind of situation requires acceptance levels at or below .5. With not purely probabilistic acceptance, this situation is more natural.

Note that non-nested $\Pi$'s would seem irrational via a Dutch Book argument, but the agent still posts consistent odds whenever he considers two or more lotteries simultaneously. It's only when he posts odds independently and they are subsequently collected that leads to inconsistency. Consult the Ellsberg paradox for intuitions here.

## IX. References

[Bar81] Barnett, J. "Computational Methods for a Mathematical Theory of Evidence," IJCAI 7.

[Che83] Cheeseman, P. "A Method of Computing Generalized Bayesian Probability Values for Expert Systems," IJCAI 8.

[Dem68] Dempster, A. "A Generalization of Bayesian Inference." *J. Roy. Stat. Soc. 2*

[Dil82] Dillard, R. "The Dempster-Shafer Theory Applied to Tactical Fusion in an Inference System," Fifth MIT/ONR Workshop.

[Fis65] Fishburn, P. "Analysis of Decisions with Incomplete Knowledge of Probabilities," *Op. Res. 13*.

[GLF81] Garvey, T., Lowrance, J., and Fischler, M. "An Inference Technique for Integrating Knowledge from Disparate Sources," IJCAI 7.

[Gin84] Ginsberg, M. "Non-monotonic Reasoning using Dempster's Rule," AAAI.

[Goo83] Good, I. *Good Thinking: The Foundations of Probability and Its Applications*, Minnesota.

[GoM84] Good, I. and McMichael, A. "A Pragmatic Modification of Explicativity for the Acceptance of Hypotheses," *Phil. Sci. 51*.

[Hur51] Hurwicz, L. "Some Specification Problems and Applications to Econometric Models," *Econometrica 19*

[Kyb70] Kyburg, H. "Conjunctivitis," in Swain, M. ed., *Induction, Acceptance, and Rational Belief*, Dordrecht: Reidel.

[Kyb74] Kyburg, H. *The Logical Foundations of Statistical Inference*, Dordrecht: Reidel.

[Kyb85] Kyburg, H. *Science and Reason*, manuscript.

[Lev80a] Levi, I. "Acceptance as a Basis for Induction," in Cohen and Hesse, eds., *Applications of Inductive Logic*, Oxford.

[Lev80b] Levi, I. *The Enterprise of Knowledge*, MIT, Cambridge

[Lop84] Lopes, L. "Normative Theories of Rationality: Occam's Razor, Procrustes' Bed?" (response to Kyburg), *Behav. Brain Sci. 6*.

[Lou84] Loui, R. "Primer for the Inexact Reasoning Session," memorandum.

[Lou85] Loui, R. "Decisions with Indeterminate Probabilities," submitted.

[Low82] Lowrance, J. "Dependency-Graph Models of Evidential Support," Ph.D. Thesis, U. Mass. Amherst, COINS TR.

[LuS84] Lu, S. and Stephanou, H. "A Set-theoretic Framework for the Processing of Uncertain Knowledge," AAAI.

[Nil84] Nilsson, N. "Probabilistic Logic" SRI Tech. Note 321

[Qui83] Quinlan, J. "Consistency and Plausible Reasoning," IJCAI 8.

[Sha76] Shafer, G. *A Mathematical Theory of Evidence*. Princeton.

[Sta66] Starr, M. "A Discussion of Some Normative Criteria for Decision-Making under Uncertainty," *Industrial Management Review 8*.

[Str84] Strat, T. "Continuous Belief Functions for Evidential Reasoning," AAAI.

[TSD83] Tong, R., Shapiro, D., Dean, J., and McCune, B. "A Comparison of Uncertainty Calculi in an Expert System for Information Retrieval," IJCAI 8.

[WeH82] Wesley, L. and Hanson, A. "The Use of an Evidential Based Model for Representing Knowledge and Reasoning about Images in the Vision System," *IEEE*.

[Wes83] Wesley, L. "Reasoning about Control: The Investigation of an Evidential Approach," IJCAI 8.

[WLG84] Wesley, L., Lowrance, J., and Garvey, T. "Reasoning about Control: An Evidential Approach," SRI Tech. Note 324.

[Zad79] Zadeh, L. "On the Validity of Dempster's Rule," UC Berkeley/ERL M79/24.